\documentclass[runningheads]{llncs}
\usepackage{makecell}
\usepackage{graphicx}
\usepackage[utf8]{inputenc} 
\usepackage[T1]{fontenc}    
\usepackage{url}            
\usepackage{nicefrac}       
\usepackage{microtype}      
\usepackage{xcolor}         
\usepackage{bm}
\usepackage{bbm}
\usepackage{todonotes}
\usepackage{algorithm}
\usepackage{algpseudocode}
\usepackage{multicol}
\usepackage{caption}
\usepackage{multirow}
\usepackage{makecell}
\usepackage{booktabs} 
\usepackage{amsfonts} 
\usepackage{graphicx} 
\usepackage{duckuments} 
\usepackage{bold-extra}
\usepackage{amsmath}
\usepackage[square,sort,numbers]{natbib}  

\newcommand{\R}{\mathbb{R}}

\newcommand{\X}{X}

\newcommand{\Loss}{\mathcal{L}}

\newcommand{\inputs}{\mathcal{X}}
\newcommand{\outputs}{\mathcal{Y}}

\newcommand{\weights}{\boldsymbol{\omega}}
\newcommand{\forget}{\mathcal{D}_f}
\newcommand{\remain}{\mathcal{D}_r}
\newcommand{\ori}{h_{\weights}}
\newcommand{\un}{h_{\weights^{u}}}
\newcommand{\optimal}{h_{\weights^{*}}}

\newcommand{\cfinst}{\X_{j}^{cf}}
\newcommand{\SCM}{\mathcal{G}}
\newcommand{\argmin}{\text{arg\,min}}

\begin{document}
\title{Debiasing Machine Unlearning with Counterfactual Examples\thanks{Supported by organization x.}}


\author{Ziheng Chen\inst{1,3} \and
Jia Wang\inst{2} \and
Jun Zhuang\inst{4} \and Abbavaram Gowtham Reddy\inst{5} \and 
Fabrizio Silvestri\inst{6} \and Jin Huang\inst{1} \and Kaushiki Nag\inst{3} \and Kun Kuang\inst{7} \and Xin Ning\inst{8} \and Gabriele Tolomei\inst{6}}
\authorrunning{Z. Chen et al.}
%
\institute{Applied Mathematics and Statistics Department, Stony Brook University \and The Xi’an Jiaotong-Liverpool University \and Walmart Global Tech \and Computer Science Department, Boise State University \and Indian Institute of Technology Hyderabad \and Sapienza University of Rome, Italy \and Zhejiang University \and Institute of semiconductors, Chinese Academy of Sciences}

\maketitle 
\begin{abstract}
 The right to be forgotten (RTBF) seeks to safeguard individuals from the enduring effects of their historical actions by implementing machine-learning techniques. These techniques facilitate the deletion of previously acquired knowledge without requiring extensive model retraining. However, they often overlook a critical issue: unlearning processes bias. This bias emerges from two main sources: (1) data-level bias, characterized by uneven data removal, and (2) algorithm-level bias, which leads to the contamination of the remaining dataset, thereby degrading model accuracy. In this work, we analyze the causal factors behind the unlearning process and mitigate biases at both data and algorithmic levels. Typically, we introduce an intervention-based approach, where knowledge to forget is erased with a debiased dataset. Besides, we guide the forgetting procedure by leveraging counterfactual examples, as they maintain semantic data consistency without hurting performance on the remaining dataset. Experimental results demonstrate that our method outperforms existing machine unlearning baselines on evaluation metrics.

\end{abstract}

\section{Introduction}
\label{sec:intro}

The rise of data protection regulations, notably the \textit{``right to be forgotten,''} has spurred the need for technologies to selectively erase knowledge acquired while training machine learning/artificial intelligence (ML/AI) models without requiring full retraining. These regulations empower individuals to request personal data removal from ML/AI systems, ensuring privacy compliance. In response, \textit{machine unlearning} techniques have become vital for efficiently discarding specific model knowledge and addressing data privacy and compliance requirements.

However, the development of machine unlearning methods has overlooked a fundamental challenge: the potential introduction of bias stemming from non-uniform data removal during the unlearning process~\cite{zhang2024forgotten,oesterling2023fair,kadhe2023fairsisa}. Moreover, recent methods~\cite{wu2022puma,kurmanji2023towards} aim to increase error for forgotten data points, potentially assigning them to semantically inconsistent class or even random class~\cite{chundawat2023can}. Such methods can contaminate the remaining data, resulting in the creation of bias and performance degradation~\cite{zhang2024forgotten, oesterling2023fair}.

\begin{figure}[t]
\label{fig:Boundary}
\centering
\includegraphics[width=0.5\columnwidth]{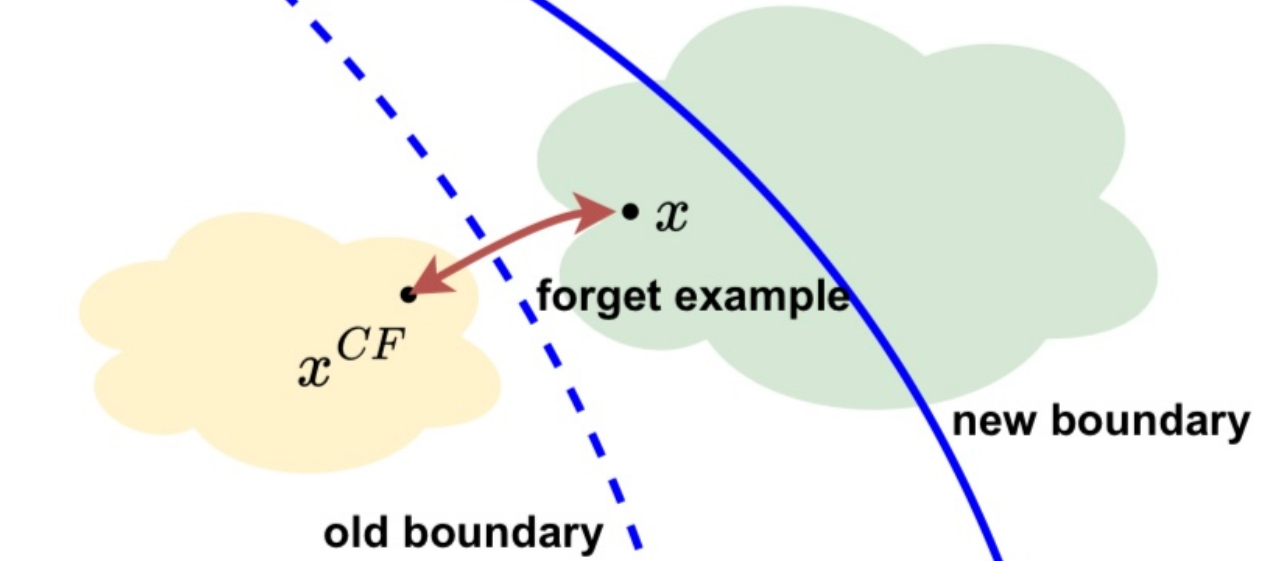}
\caption{Expanding old decision boundaries to erase causal information associated with forgotten examples. The green and yellow areas denote two distinct yet semantically closed classes.}
\label{fig:example}
\vspace{-0.4cm}
\end{figure}

To mitigate bias during the unlearning process, we examined the impact of retraining an unbiased model without including the samples to be forgotten. Our analysis revealed that such retraining can lead to misclassifying forgotten examples into semantically related categories. Specifically, in the \textit{CIFAR-100} dataset, forgetting instances of the ``rabbit'' class resulted in approximately 82\% of those samples being incorrectly classified as either the ``hamster'', ``mouse'', or ``squirrel'' classes (37\%, 26\%,  and 19\% respectively), which are closely related felines. Among these images, 73\% are classified into the same class as their \textit{counterfactual examples} (CFs) generated by Semantic Consistent Visual~\cite{vandenhende2022making}. 
A counterfactual example for a given instance is generated by applying the minimal perturbation to the input that alters the prediction, causing the resulting modified sample to cross the decision boundary induced by the model.
CFs have been successfully used to attach post-hoc explanations for predictions in the form: ``\textit{If A had been different, B would \textbf{not} have occurred}''~\cite{stepin2021survey}.

The finding above inspires us to guide the model to forget. We leverage CFs as pivotal points to encompass forgotten samples into semantically similar classes. As illustrated in Figure 1, our approach aims to make forgotten samples and their counterfactuals (CFs) indistinguishable by the model, effectively broadening the local decision boundary. This strategy minimizes the impact of forgetting on the adjacent remaining samples.

In this work, we adopt a causal perspective~\cite{ruan2022causal} to examine bias during the unlearning process and address data-level bias through causal intervention. Subsequently, we investigate algorithmic bias and employ counterfactual examples as a mitigation strategy. Our main contributions are summarized as follows:\\
$(i)$ We propose a causal framework to formulate the machine unlearning procedure and analyze the potential source of bias induced \textit{(Section~\ref{sec:scm})}. \\
$(ii)$ We introduce an intervention-based approach to alleviate bias-induced impacts on remembering caused by selective sample deletion during unlearning \textit{(Section~\ref{sec:eradata})}. \\
$(iii)$ We guide the forgetting procedure leveraging \textit{counterfactual examples}, as they maintain semantic data consistency without hurting remaining samples \textit{(Section~\ref{sec:eraalg})}. \\
$(iv)$ We validate our approach in both uniform and non-uniform deletion setups \textit{(Section~\ref{sec:experiments})}.

\section{Related Work}
\label{sec:related}

\noindent \textbf{Machine Unlearning.}
To ensure security and data privacy, it's vital to enable dataset or model modifications without retraining, termed \textit{machine unlearning}. Most existing works can be categorized into model-agnostic approaches, model-intrinsic approaches, and data-based approaches~\cite{nguyen2022survey,cui2023variational,li2022revisiting}. Model-agnostic approaches involves statistical query learning~\cite{cao2015towards}, decremental learning~\cite{chen2019novel}, and knowledge adaptation~\cite{chundawat2023can, zhao2023stable}. Meanwhile, model-intrinsic approaches include unlearning methods designed for specific types of models. Conversely, data-based approaches encompass strategies such as data augmentation~\cite{huang2021unlearnable, liu2024please}, and data impact~\cite{peste2021ssse}. Our method improves upon existing machine unlearning techniques by addressing the bias issue inherent in the unlearning process.

\noindent \textbf{Bias in Machine Unlearning.} Bias is a common issue in ML/AI models as the model may output unfair results given that the model is trained with a biased dataset. This problem has been widely studied in recent years~\cite{ahmad2021s}. Most existing studies aim to address the bias issue at the group level~\cite{hardt2016equality} and the individual level~\cite{udeshi2018automated}.
To address the bias issue in machine unlearning, Zhang et al. first evaluate machine unlearning methods to assess the de-bias implications~\cite{zhang2024forgotten, mao2023less}. Oesterling et al. propose a new machine unlearning method to eliminate group bias~\cite{oesterling2023fair}. Our method introduces a novel approach by utilizing counterfactual examples to guide the debiasing procedure, thereby ensuring semantic data consistency to effectively mitigate bias. This innovation leads to a more robust solution to the challenges encountered in machine unlearning.

\noindent \textbf{Counterfactual Explanations.}
Counterfactual Examples offer post-hoc explanations for predictions through statements like "If A had been different, B would not have occurred", known as counterfactual explanations. These explanations have been applied in various domains, including natural language understanding~\cite{tian2022debiasing, chen2023explainable}, data augmentation~\cite{qiang2022counterfactual}, and medical applications~\cite{abid2022meaningfully}. Methods for generating counterfactual explanations can be categorized into model-specific approaches, which tailor machine learning models~\cite{lucic2022focus, gao2024dlora}, and model-agnostic approaches, which can provide explanations for different models~\cite{guidotti2018local, zhou2023xai,guan2024ufid}.

\section{Problem Formulation}
\label{sec:background}

Let $\inputs \subseteq \R^n$ be an input feature space and $\outputs$ an output label space. Without loss of generality, we consider the $K$-ary {\em classification} setting, i.e., $\outputs = \{0, \ldots, K-1\}$. The training set can be represented as $\mathcal{D}=\bigcup\limits_{i\in \{0, \ldots, K-1\}}\mathcal{D}_i$ where $\mathcal{D}_i$ contains data points corresponding to class $i$.  We introduce a predictive model $\ori: \inputs \mapsto \outputs$, with parameters $\weights$, trained on the entire dataset $\mathcal{D}$ by minimizing the loss function $\mathcal{L}_{(X_i,Y_i) \in \mathcal{D}}(\ori(X_i),Y_i)$. We refer to $\ori$ as the ``original model,'' which maps input feature vectors $X \in \inputs$ to predictive labels $\ori(X) = \hat{Y} \in \outputs$.

In this work, we focus on the problem of deep machine unlearning for classification. This task involves selectively removing specific instances or knowledge from a trained model without retraining the entire model. We denote the set of samples to forget as $\mathcal{D}_f \subset \mathcal{D}$. 
The set of samples to be retained is denoted as $\remain$, with the assumption that $\mathcal{D}_f$ is the complement of $\remain$, i.e., $\mathcal{D}_f \cup \remain = \mathcal{D}$ and $\mathcal{D}_f\cap \mathcal{D}_r = \emptyset$. The goal of machine unlearning is to obtain a new model $\un$ that removes the information contained in $\mathcal{D}_f$ from $\ori$ without negatively impacting its performance on $\remain$. Given the time-consuming nature of fully retraining the model on $\remain$ to obtain an expected model $\optimal$, our target is to approximate $\optimal$ by evolving $\ori$ using $\mathcal{D}_f$ as follows:
\vspace{-0.3cm}
\[
\ori \xrightarrow{\forget} \un \simeq \optimal
\]

\section{Structural Causal Modeling for Classification}
\label{sec:scm}
\vspace{-1.1cm}
\begin{figure}[h]
\centering
\includegraphics[width=0.5\columnwidth]{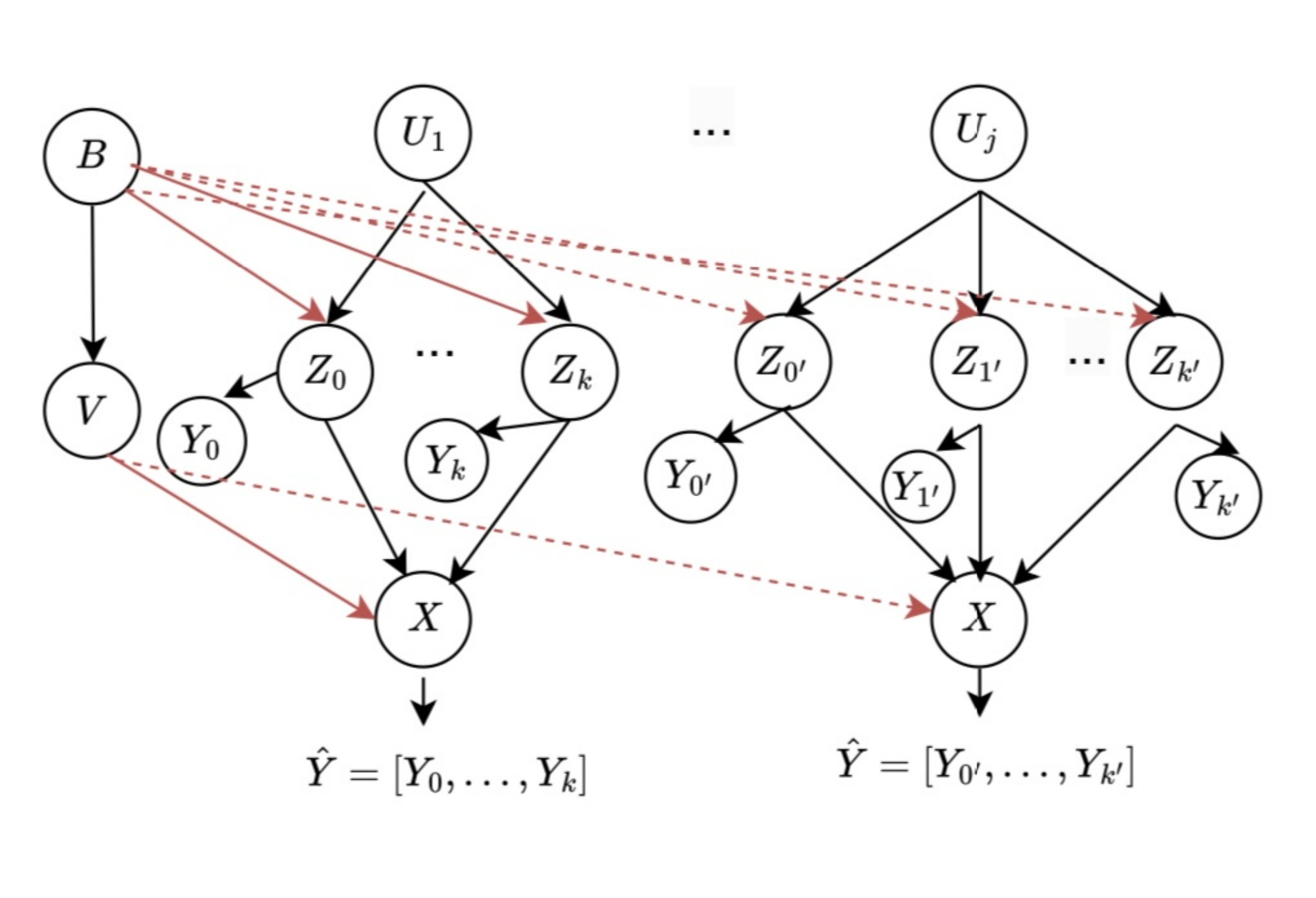}
\vspace{-0.6cm}
\caption{Overview of the SCM. We denote $X$, $\mathcal{Z}$, $Y$, $V$, $B$, and $U$ as real-world examples, causal factors, classes, background variables, domain variables, and class-level concepts, respectively.}
\label{fig:1}
\vspace{-0.6cm}
\end{figure}

To understand the emergence of bias during unlearning, we start by formulating a Structural Causal Model (SCM) $\SCM$~\cite{pearl2009causal, chu2024task}, which captures the underlying causal generative processes of the original model $\ori$ as depicted in Fig. \ref{fig:1}~\cite{ruan2022learning, liu2021neural}. Generally, we categorize the latent factors responsible for generating real-world examples $X$ into two groups: background variables $V$ (e.g., image background, style, etc.), controlled by a domain variable $B$, and causal factors $\mathcal{Z}=\bigcup_{i=0}^{K-1} Z_i$. Each $Z_i$ corresponds to a specific class $Y_i$ (e.g., $Z_i$ represents the causal factor for rabbits).
Importantly, these causal factors $\mathcal{Z}$ can be grouped into $m$ underlying class-level concepts $\mathcal{U}=\bigcup_{j=1}^{m} U_j$, and each $Z_i$ has only one class-level concept as its parent: $Pa_{Z_i}={U_j | U_j\rightarrow Z_i}$. For example, both $Z_i$ representing rabbits and $Z_j$ representing mice can be traced back to the common class-level concept root $U_j=Pa_{Z_j}=Pa_{Z_i}$ for rodents. Due to similarities within the same class-level concept, the data can be seen as a mixture of causal factors within the same group, interacting with the background variable $V$ (e.g., some rabbit images closely resemble other rodents like mice and squirrels). Formally, for any data point $(X_i, Y_i)\in \mathcal{D}$, the generative model for data $X_i$ can be expressed as follows:

\[
\mathcal{U} \sim P_{U},\quad B \sim P_{B},\quad Z_i=f_{i}(Pa_{Z_i},B),\]
\[X=g(Z_i, \mathcal{Z}_{-i}, V),\quad Y_i=l(Z_i),
\]
where $\mathcal{Z}_{-i}=\{ Z_j | Pa_{Z_j} = Pa_{Z_i} \mbox{and} j\neq i \}$ denotes all the causal factors that share the same parent with $Z_i$.

\subsection{The Emerging of Bias during Unlearning}
Let $\SCM$ be the SCM of an underlying causal model. The presence of confounding variables~\footnote{We claim that $X$ and $Y$ are confounded some other confounding variable $Z$ whenever $Z$ causally influences both $X$ and $Y$.} $\mathcal{U}$ and $B$ poses a risk of amplifying the following bias through spurious causal correlations during the unlearning procedure.

\begin{itemize}
\item Data-level shortcut bias~\cite{geirhos2020shortcut} via backdoor $Z_i \leftarrow B \rightarrow V \rightarrow X \rightarrow \hat{Y_i}$. The selection of $\mathcal{D}_f$ can lead to a distribution mismatch in the background variable $V$ between $\mathcal{D}_r$ and $\mathcal{D}$, which may guide $\un$ to solely capture the non-invariant spurious correlation between $Y_i$ and $V$.

\textbf{Illustrative example:}\textit{ Suppose the original dataset contains images of various animals in different environments, including rabbits against diverse backdrops. After the unlearning process, if the remaining dataset becomes biased towards rabbit instances in a specific background (e.g., grass), the unlearning model may incorrectly infer a spurious correlation between the presence of the grass background and the absence of the rabbit, leading to biased model updates.}
\vspace{0.2 cm}

\item Data-level label bias~\cite{fabbrizzi2022survey} via backdoor $Z_i \leftarrow Pa_{Z_i} \rightarrow Z_j \rightarrow X \rightarrow \hat{Y_i}$. Once there exists a dominating class in $\remain$, $\un$ tends to enhance the spurious correlation between other images and the majority class.

\textbf{Illustrative example:}\textit{
Assume that the dataset comprises an equal number of images for various species like rabbits, mice, and squirrels, representing the biodiversity accurately. Here, we consider the scenario where the unlearning process focuses excessively on eliminating images of squirrels due to certain biases or errors. As a consequence, the remaining dataset might become skewed, with fewer squirrel images compared to rabbits and mice. Consequently, the model updates during unlearning may inadvertently amplify the correlations between the remaining images and the prevalent categories.}

\vspace{0.2 cm}

\item Algorithmic bias~\cite{fabbrizzi2022survey, he2024hierarchical} is raised by unlearning algorithms targeting on increasing the classification error on $\forget$~\cite{chundawat2023can,kurmanji2023towards,chen2023boundary}. We will elaborate on this in section~\ref{sec:eraalg}.
\end{itemize}

We now propose a method to address the above biases in machine unlearning.

\section{Proposed Method}
\label{sec:method}
\begin{figure}[h]
\centering
\includegraphics[width=.75\columnwidth]{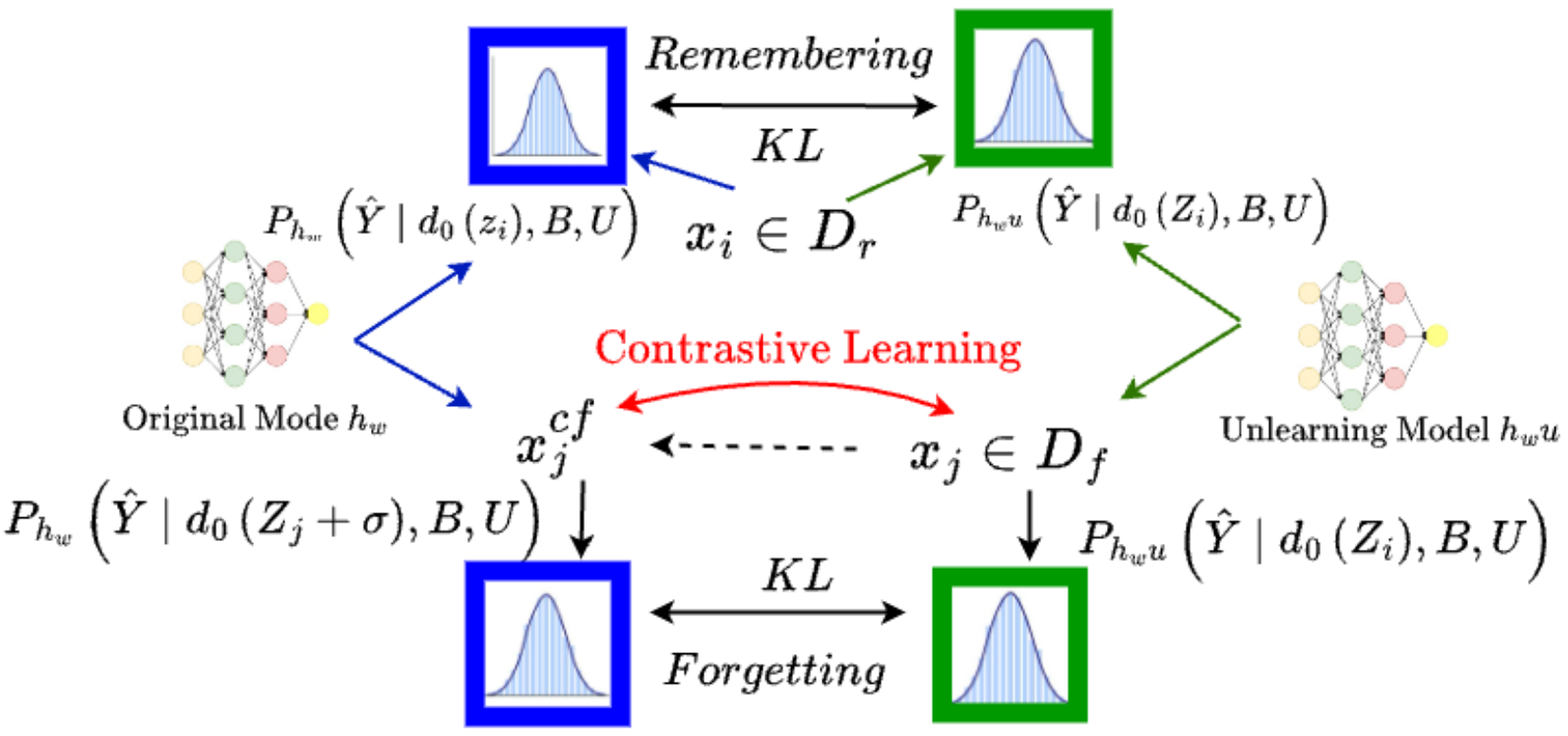}
\caption{The overall system framework.}
\label{fig:framework}
\vspace{-0.3cm}
\end{figure}

In this section, we introduce a teacher-student framework for unlearning. As shown in Fig~\ref{fig:framework}, we adopt the original model $\ori$ as the ``teacher'', and our primary objective is to train a ``student'' $\un$ capable of selectively adhering to the teacher's guidance on $\remain$ (remember)  while deliberately deviating on $\forget$ (forget). To address potential bias that may arise during this process, we incorporate causal interventions as a means of bias mitigation. 

\vspace{-0.1cm}
\vspace{-0.1cm}

\subsection{Erasing Data-level Bias in Remembering}
\label{sec:eradata}
We start by guiding the student $\un$ to obey the performance of teacher $\ori$ on $\remain$ and initialize the parameters of $\un$ with the parameters of $\ori$. Meanwhile, to erase the data-level bias, we decouple the spurious causal correlation by directly intervening on causal factors. Specifically, for a sample $X_i$ from class $Y_i$,  we adopt $P(\hat{Y}|do(Z_i))$ rather than $P(\hat{Y}|X_i)$ for prediction. This approach, according to do-calculus, cuts off the path $V \leftarrow B \rightarrow Z_i$ and $Z_j \leftarrow Pa_{Z_i} \rightarrow Z_i$, effectively mitigating both shortcut and label bias simultaneously. Formally, $P(\hat{Y}|do(Z_i))$ can be expressed as:
\begin{equation}
\begin{aligned}
P(\hat{Y}|Z_i,B, U) &= \sum_{v}P(V|B,Z_i)\sum_{\mathcal{Z}_{-i}}P(\hat{Y}|Z_i,V,\mathcal{Z}_{-i})P(\mathcal{Z}_{-i}|U, Z_i) \\
P(\hat{Y}|do(Z_i),B,U) &= \sum_{v}P(V=v|B)\sum_{z_{-i}}P(\hat{Y}|z_i,v,z_{-i})P(\mathcal{Z}_{-i}=z_{-i}|U)\\
&\approx P(\hat{Y}|z_i,\sum_{v}P(v|B)v,\sum_{z_{-i}}P(z_{-i}|U)z_{-i})
\end{aligned}
\end{equation}

In the last step, we adopt the Normalized Weighted Geometric
Mean (NWGM) approximation~\cite{kim2019bridging} to move the outer sampling over $V$ and $\mathcal{Z}_{-i}$ into the feature level. 
Driven by the observation that loss gradients from robust models exhibit better alignment with salient data features and human perception, which well outlines the contour of an object
in images~\cite{kim2019bridging, ren2022dice, mao2022accelerating}, we use the magnitude of gradients to extract $V$ and $Z_i$. Following~\cite{ren2022dice}, given an instance $X_i$ with label $Y_i$, we obtain a mask $M$, to separate $V$ from $Z_i$ as follows:
\[\delta=\nabla \Loss(\ori(X_i), Y_i),\quad M=I(\delta \leq d) \quad v=M\odot X_i, \quad z_i=(1-M)\odot X_i,\] where $d$ serves as the threshold and we select the value that excludes $50\%$ features for $v$ as d in all experiments.
Notice that, for tabular data, we also include sensitive attributes (e.g., age, gender) into $V$. 
For $P(v|B)$ and $P(z_{-i}|U)$, to avoid dependence of $D_r$, we assume they follow the uniform distribution. Hence, we have
$P(\hat{Y}|do(Z_i),B,U)  \approx P(\hat{Y}|z_i+\bar{v}+\gamma*\bar{z}_{-i})$. 
Where $\bar{v}$ and $\bar{z}_{-i}$ could be obtained via random sampling and averaging. Notice that $\bar{z}_{-i}$ could only be sampled within an image generated by causal concept $Pa_{Z_i}$.
Besides, we set $\gamma$ to 0.2 for all experiments to ensure the intervention doesn't adversely affect the instance.

Finally,  we train $\un$ to unbiasedly maintain the performance of  $\ori$  on $D_r$  by minimizing the interventional distribution as follows:
\begin{equation}
\Loss_r=\sum_{(X_i,Y_i)\in \remain}KL(P_{\ori}(\hat{Y}|do(Z_i),B,U),P_{\un}(\hat{Y}|do(Z_i),B,U))
\end{equation}

\subsection{Erasing Algorithm-level Bias in Forgetting}
\label{sec:eraalg}
\subsubsection{ Analyzing the causal reasons for algorithm-level bias.}
\begin{figure}[h]
\centering
\includegraphics[width=0.75\columnwidth]{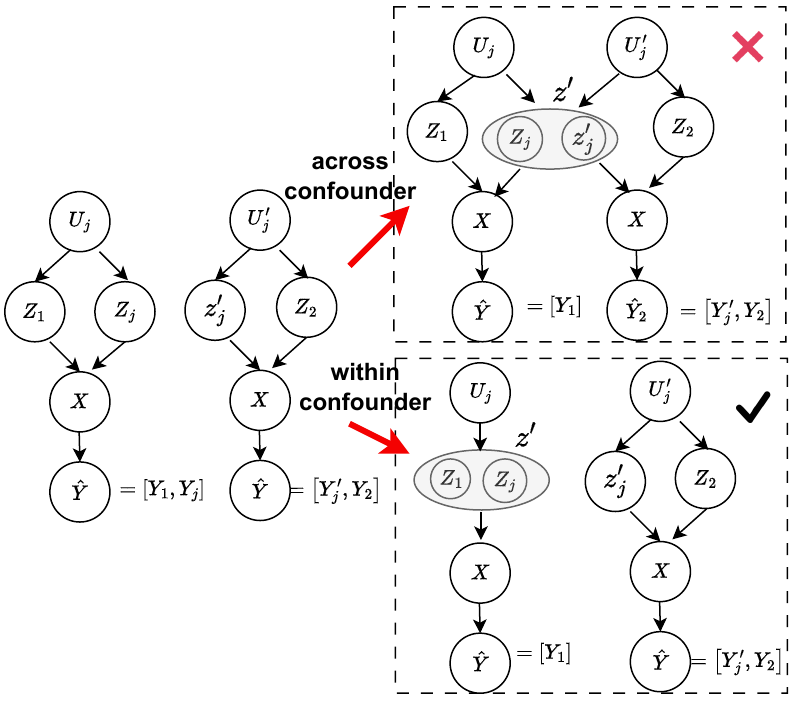}
\caption{Causal Mechanism of Forgetting.}
\label{fig:forget}
\vspace{-0.3cm}
\end{figure}

To erase information contained in $\forget$, existing methods~\cite{chundawat2023can,chen2023boundary,kurmanji2024towards,wu2022puma} typically focus on increasing the classification error on the subset $\forget$. This involves deliberately assigning incorrect labels to forget samples $(X_j, Y_j)\in \forget$~\cite{chundawat2023can, ma2022learning}. However, this strategy poses a potential risk of affecting the performance of $\un$ on $\remain$, as it introduces spurious correlations and undermines the causal relationships learned by $\ori$. For instance, instructing $\un$ to forget rabbit images by misclassifying them as eagles might compel it to rely on shortcut features shared between rodents and birds images, despite their significant differences. This is illustrated in Fig. ~\ref{fig:forget} (across confounder), where assigning incorrect labels ($Y^{'}_j$) from disparate class-level concepts ($ U^{'}_j \neq U_j $) to forget examples ($Y_j$) can create new confounder pairs ($Z_j \leftrightarrow U^{'}_j \leftrightarrow Z_2$). 

Based on this insight, assigning $X_j$ with a label of a similar class concept is critical, as demonstrated in Fig.~\ref{fig:forget} (within confounder). In line with the example above, a strategy for $\un$ to ``forget'' rabbits might involve classifying them as mice, thus emphasizing common rodent features and ignoring rabbit-specific characteristics.
\vspace{-4mm}
\subsubsection{Debias through counterfactual examples.}
To mitigate algorithmic bias, we employ counterfactual examples (CFs) that maintain semantic consistency while reversing the prediction. These CFs act as anchors, enabling the integration of forgotten samples into classes within the same class-level concept.

In explainable machine learning, the focus has been on generating CFs that shift to a different class with minimal changes to causal features. Generally, given an instance $X_j$ its optimal CF is defined as $X_j^{cf} = X_j + \delta^*$ and can be found  by solving the following constrained objective:
\begin{equation}
\begin{aligned}
\delta^*~=~\argmin_{\delta\in Z_j}(X_j, X_j + \delta)\\
\textrm{subject to:} \quad \ori(X_j)\neq\ori(\cfinst)
\end{aligned}
\end{equation}

Here $\delta$ represents the optimal action guiding $\cfinst$, and the magnitude function $M$ discourages $\cfinst$ from being too far away from $\X_j$.
Crucially, the condition $\delta \in Z_j$ guarantees that $\cfinst$ is formulated by merely altering causal-relevant features and maintaining the remains. To achieve this, we adopt the following algorithms for different types of forgotten data:
\begin{itemize}

\item {\bf \em Headtails}~\cite{vandenhende2022making}: 
 For image data, the CF of one sample is generated by only replacing the semantically matching image regions from multiple distractor images sharing the same class-level concept. To flip the label with minimal alteration, the method involves segmenting images into distinct cells and identifying the most discriminative section for replacement.
\item {\bf \em C-CHVAE}~\cite{pawelczyk2020learning}: For tabular data, it adopts a variational autoencoder as a search device to find CFs that are proximate and connected to the input data. Hence, the CFs will be classified into another 
semantically similar category relative to the input data.
\end{itemize}

\subsubsection{Unbiased Forgetting by Distribution Alignment}

With $\cfinst$, we aim at guiding $\un$ to forget the distinct features of $X_j$ and treat it as the same class as $\cfinst$. Naturally, we minimize the KL divergence to align the interventional distribution of $X_j$ closely with that of $\cfinst$:
\begin{equation}
\label{forget}
\Loss_f=\sum_{(X_j,Y_j)\in \forget}KL(P_{\ori}(\hat{Y}|do(Z_j+\delta),B,U),P_{\un}(\hat{Y}|do(Z_j),B,U))
\end{equation}

\vspace{-0.2cm}
\subsubsection{Unbiased Forgetting by Contrastive Learning}  
Although minimizing Loss~\ref{forget} may potentially lead the model $\un$ to treat the forget sample and its CFs indistinguishably, it does not guarantee their geometric proximity within the representation space. Inspired by~\cite{tang2022contrastive, ma2024data}, our goal is to reduce the distance between $X_j \in \forget$ and $\cfinst$, thereby allowing the decision boundary to expand slightly to encompass $X_j$.  From this intuition, we aim to contrast each forget sample with 1) the corresponding counterfactual example (positive pairs) to pull their representations closer, and 2) remaining samples from distinct classes (negative pairs) and push their representation apart. 

Concretely, we denote the negative sample set as $N(X_j)=\{X^{-}_k|X^{-}_k \in \remain, \ori(X^{-}_k) \neq \ori(X_j)\}$. For simplicity, the representation of $X_j$, $\cfinst$ and $X^{-}_k$ learned by $\un$ could be denoted as $\xi_j$, $\xi_j^{cf}$ and $\xi_k^{-}$ respectively.

The contrastive boundary loss aims to maximize the similarity of positive pairs and minimize the similarity of negative pairs as follows:
\[\Loss_{f}^{CB}=\sum_{X_j \in \forget} \frac{-1}{|N(X_j)|}\mbox{log}\frac{\mbox{exp}(\xi_j \cdot \xi_j^{cf}/\tau)}{\mbox{exp}(\xi_j \cdot \xi_j^{cf}/\tau)+\sum\limits_{X_k^{-} \in N(X_j)} \mbox{exp}(\xi_j \cdot \xi_k^{-}/\tau)},\]
where $\tau$ is a temperature parameter. In this part, our analysis demonstrates that optimizing $\Loss_f^{CB}$ exerts a small impact on $\remain$. This conclusion is supported by an analysis of the gradient. 

\begin{proposition}
As shown in the Supplementary, the gradient for $\Loss^{CB}_{f}$ with respect to the embedding $\xi_k^{-}$ has the following form:
\[\nabla_{\xi_k^{-}}\Loss^{CB}_{f} = \frac{\mbox{exp}(\xi_j \cdot \xi_k^{-}/\tau)}{M}\cdot\frac{\xi_j}{\tau},\]

\mbox{where} $M=\mbox{exp}(\xi_j \cdot \xi_j^{cf}/\tau)+\sum\limits_{X^{-}_k \in N(X_j)} \mbox{exp}(\xi_j \cdot \xi_k^{-})/\tau)$.
\end{proposition}

The proposition shows that the gradient aligns with the direction of $\xi_j$, and the magnitude of $\nabla_{\xi_k^{-}}\Loss^{CB}_{f}$ is contingent on the similarity between $\xi_j$ and $\xi_k^{-}$. According to~\cite{das2019separability}, 
the model optimized with cross-entropy loss produces lower geometric similarity among embeddings of samples from different classes. This implies that the product of $\xi_j \dot \xi_k^{-}$ is expected to be minimal. Thus, it is plausible to conclude that updating  $\un$ via the optimization of $\Loss^{CB}$ does not significantly affect $\remain$.

\subsection{Overall Training for Unlearning }
Finally, we get the unlearning model $\un$ by minimizing the $\mathcal{L}$:
\[\mathcal{L}=\Loss_r+\alpha \Loss_f+\beta \Loss_{f}^{CB}+\gamma\sum_{(X_i, Y_i)\in \remain}\ell(X_i,Y_i), \]
where $\ell$ stands for the cross-entropy loss, and we denote $\alpha$, $\beta$ and $\gamma$ as balanced hyper-parameters.

\section{Experiments}
\label{sec:experiments}
To evaluate the effectiveness  of our proposed model in addressing the following research questions, we designed a series of experiments:
\begin{itemize}
    \item \textbf{Q1}: How does the overall performance of our method compare with existing unlearning algorithms under a uniform deletion strategy?
    \item \textbf{Q2}: How does our algorithm perform under a non-uniform deletion strategy? Moreover, does it mitigate shortcut and label biases in scenarios of selective forgetting?
    \item \textbf{Q3}: How to determine the contribution of each loss function component to the overall model performance?
\end{itemize}
\vspace{-6mm}
\subsection{Experimental Setup}

In this section, we present our experimental findings. Initially, we assess the accuracy and efficiency of our method under random data deletion. Then, following~\cite{zhang2024forgotten}, we explore the impact of non-uniform data deletion on bias reduction.

\noindent{\bf Datasets and Models}
Our proposed method is evaluated across diverse datasets, including CIFAR-100~\cite{chundawat2023can, xiang2023tkil,liu2023riatig} and CUB200~\cite{vandenhende2022making, you2022incremental} for image classification, and Adult~\cite{oesterling2023fair} and German~\cite{oesterling2023fair} datasets for tabular data classification~\cite{ma2023learning}. CIFAR-100 utilizes provided super-classes (e.g., small mammals: rabbits, mice, squirrels) while CUB-200 classes are clustered into 10 concepts based on a confusion matrix, merging frequently confused subclasses into unified super-class. Given that both tabular datasets have binary labels, we didn't consider their class-level concepts. We adopt ResNet18 and ResNet34 architectures for image classification tasks, consistent with prior work~\cite{chundawat2023can}. For tabular data unlearning, we utilized a 3-layer DNN model~\cite{wu2022puma, ye2023medlens}.

\noindent{\bf Evaluation Measures} 
\begin{itemize}
     \item \textbf{\textit{Remaining Accuracy (RA) on $\remain$}}~\cite{wu2022puma}: This denotes the accuracy of $\un$ on $\remain$, reflecting the fidelity of $\un$.
     \item \textbf{\textit{Forgetting Accuracy (FA) on $\forget$}}~\cite{wu2022puma}: This denotes the accuracy of $\un$ on $\forget$, reflecting the ability of $\un$ in full class deletion.
     \item \textbf{\textit{Membership Inference Attack (MIA) on $\forget$}}~\cite{wu2022puma}: MIA estimates the number of samples in $\forget$ correctly predicted as forgotten by a confidence-based MIA predictor.
     \item \textbf{\textit{Run-Time Efficiency (RTE)}}~\cite{wu2022puma}: RTE quantifies the run-time cost of the proposed algorithm. 
     \item \textbf{\textit{Disparate Impact (DI)}}~\cite{zhang2024forgotten}: DI measures the ratio of favorable outcomes for the unprivileged group $(x=0)$ to the privileged group $(x=1)$: $\mbox{DI}=\frac{P(\hat{y}=1|x=0)}{P(\hat{y}=1|x=1)}$. 
      \item \textbf{\textit{Equal opportunity difference(EOD) }}~\cite{zhang2024forgotten}: EOD evaluates the difference in true positive rate between unprivileged group and privileged group.
      $\mbox{EOD}=P(\hat{y}=1|x=0,y=1)-P(\hat{y}=1|x=1,y=1)$
\end{itemize}

\noindent{\bf Counterfactual Explanation Methods} As described in Sec~\ref{sec:eraalg}, we adopt \textbf{\textit{Headtails}} for image data and \textbf{\textit{C-CHVAE}} for tabular data to generate CFs respectively.

\noindent{\bf Hyperparameters} 
We conduct an ablation study to determine the optimal value of hyperparameters by measuring its performance when $\beta/\alpha \in \{0.1, 0.2, 0.6, 1.0\}$ and $\gamma/\alpha \in \{0.1, 0.2, 0.6, 1.0\}$. Finally, we determine the best combination for each task. For the ResNet18 and ResNet34, we use pretrained models. During the unlearning procedure, we run 5 epochs with a learning rate of 0.0001 for all datasets.

\noindent{\bf Baseline Models.}
We adopt the unlearning method as follows: Badteacher (BT)~\cite{chundawat2023can}, Boundary-shift (BS)~\cite{chen2023boundary}, SISA~\cite{bourtoule2021machine}, 
PUMA~\cite{wu2022puma},
SCRUB~\cite{kurmanji2023towards}. These methods achieve unlearning in an already trained model without putting any constraints. F-unlearning~\cite{oesterling2023fair} is an unbiased unlearning method designed for tabular data and we leverage it in non-uniform deletion.
Retrain refers to training from scratch on $\remain$. In the case of non-uniform deletion, we retrain the model without bias.

\subsection{Overall Performance of Unlearning Model under a Uniform Deletion Strategy (Q1)}

\begin{figure}[ht!]
\vspace{-1.1cm}
\centering
\includegraphics[width=1\columnwidth]{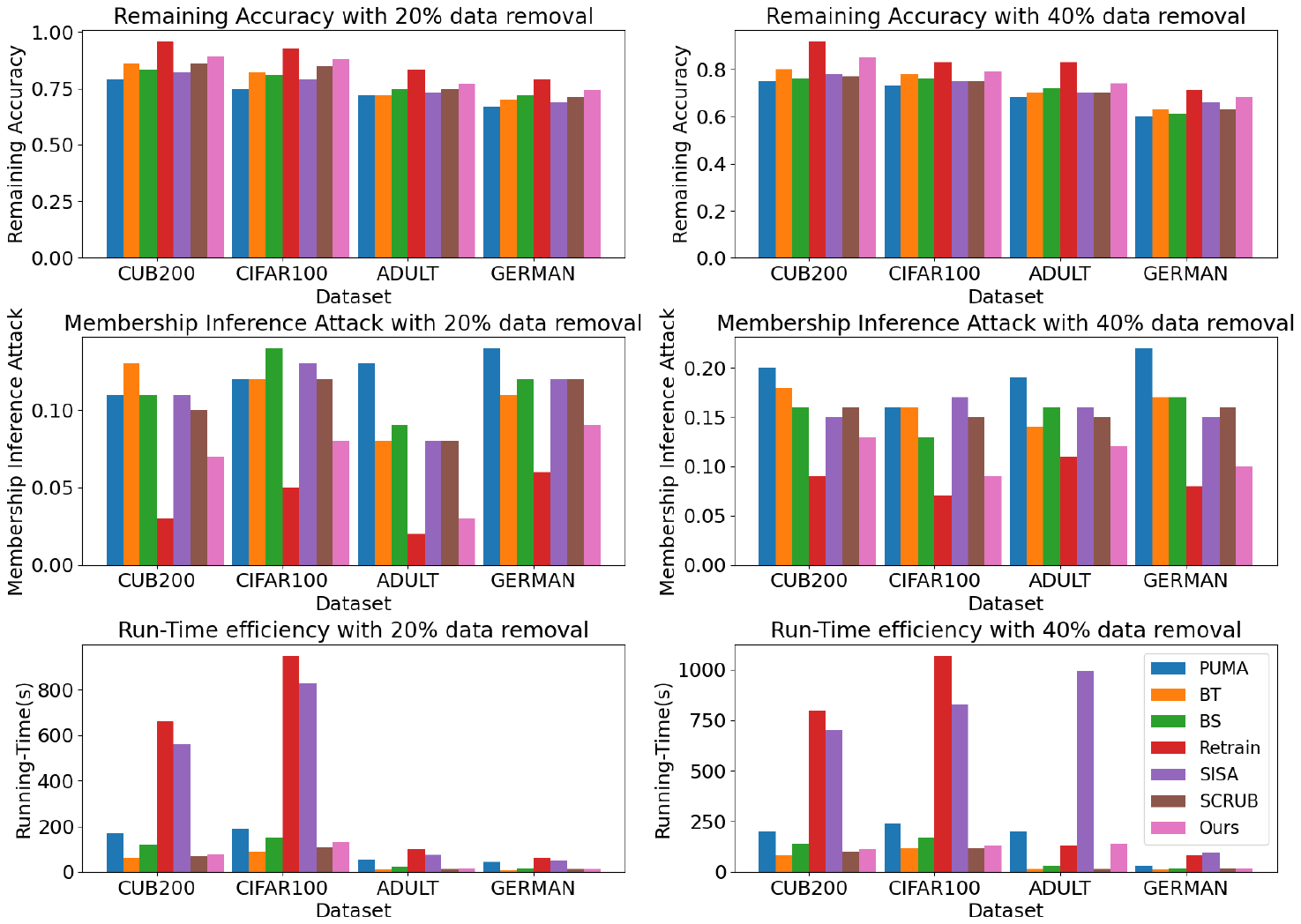}
\vspace{-0.7cm}
\caption{Result under Uniform Deletion Strategy.}
\label{fig:uniform_deletion}
\vspace{-0.8cm}
\end{figure}

\subsubsection{Sample Deletion} 
We evaluate our method's performance under uniform-sample removal by progressively eliminating training data points at percentages of 20\% and 40\% on four datasets: CUB-200, CIFAR-100, Adult, and German. This evaluation aims to demonstrate performance preservation, data removal effectiveness, and efficiency, measured by RA, MIA, and RTE, respectively. Our method consistently outperforms or performs similarly to other methods, as depicted in Figure~\ref{fig:uniform_deletion}. Notably, our approach achieves an MIA of 0.07, compared to 0.13 for the Badteacher model and 0.11 for the SCRUB. Moreover, our method maintains strong performance even with a larger proportion of forgotten data. Specifically, RA outperforms BT by 8\% and SCRUB by 5\% when 40\% of the data is removed. Although slower than Badteacher, which assigns forgetting data randomly, our method is 1.2 times faster than Boundary and 5.6 times faster than SISA.

\subsubsection{Class Deletion}

\begin{table}
\centering
\caption{ Class-level unlearning on CIFAR100 and CUB200 with ResNet18 and ResNet34. The results show the accuracy on $\remain$ and $\forget$.}
\label{tab:class-unlearning}
\scalebox{0.75}{
\begin{tabular}{c|c|c|c|c|c|c|c|c|c}
\toprule
        {\textbf{Setting}}            &      {\textbf{\# Class}}               & {\textbf{Set}}  & {\textbf{Retrain}}  & {\textbf{BT}} & {\textbf{BS}}  & {\textbf{SISA}} & {\textbf{PUMA}} & {\textbf{SCRUB}} & {\textbf{Ours}} \\
\midrule
\multirow{4}{*}{\thead{RES18+ \\C1IFAR100}} & \multirow{2}{*}{1} & $\remain$ & $ \bf 78.5 (\pm 3.2)$ & $75.5 (\pm 2.3)$ & $76.5 (\pm 2.8)$ & $76.1 (\pm 2.6)$ & $74.8 (\pm 3.2)$ & $76.3 (\pm 2.9)$ & $\bf 77.1 (\pm 2.4)$ \\ \cline{3-10} 
      &              & $\forget$ & $\bf 0.0 (\pm 0.0)$ &$4.2 (\pm 0.8)$  & $3.2 (\pm 0.6)$ & $\bf 0.0 (\pm 0.0)$ &$5.2 (\pm 0.8)$  & $\bf 0.0 (\pm 0.0)$ & $\bf 0.0 (\pm 0.0)$ \\ \cline{2-10} 
                  & \multirow{2}{*}{20} & $\remain$ & $\bf 78.2 (\pm 2.9)$ & $72.3 (\pm 2.1)$ & $75.2 (\pm 2.5)$ & $76.1 (\pm 2.3)$ & $72.5 (\pm 2.9)$ & $74.9 (\pm 2.3)$ & $\bf 76.8 (\pm 2.3)$ \\ \cline{3-10} 
                  &                   & $\forget$ &$\bf 0.0 (\pm 0.0)$  & $6.9 (\pm 0.7)$ & $6.1 (\pm 0.5)$ & $\bf 5.6 (\pm 0.6)$ & $8.3 (\pm 0.8)$ & $6.3 (\pm 0.5)$ & $\bf 5.6 (\pm 0.6)$ \\ \hline
\multirow{4}{*}{\thead{RES18+ \\CUB200}} & \multirow{2}{*}{1} & $\remain$ & $\bf 83.0 (\pm 4.3)$ & $76.6 (\pm 4.7)$ & $77.1 (\pm 4.3)$ & $78.3 (\pm 3.9)$  & $78.1 (\pm 4.9)$ &$78.9 (\pm 4.3)$  & $\bf 81.6 (\pm 3.9)$ \\ \cline{3-10} 
                  &                   & $\forget$ & $\bf 0.0 (\pm 0.0)$  & $3.25 (\pm 0.5)$  & $2.76.0 (\pm 0.5)$  & $\bf 0.0 (\pm 0.0)$  & $3.3 (\pm 0.6)$  & $\bf 0.0 (\pm 0.0)$  & $\bf \bf 0.0 (\pm 0.0)$  \\ \cline{2-10} 
                  & \multirow{2}{*}{20} &  $\remain$& $\bf 82.3 (\pm 3.7)$  & $75.9 (\pm 3.5)$  & $76.5 (\pm 3.5)$  & $78.7 (\pm 3.2)$  & $75.3 (\pm 3.3)$  & $78.6 (\pm 3.4)$  & $\bf 79.9 (\pm 3.2)$  \\ \cline{3-10} 
                  &                   &  $\forget$& $\bf 0.0 (\pm 0.0)$& $9.2 (\pm 0.8)$ & $8.6 (\pm 0.3)$ & $\bf 7.1 (\pm 0.6)$ & $8.6 (\pm 0.3)$ & $8.2 (\pm 0.3)$ &$7.3 (\pm 0.3)$  \\ \hline
\multirow{4}{*}{\thead{RES34+ \\CIFAR100}} & \multirow{2}{*}{1} & $\remain$ & $\bf 79.2 (\pm 3.5)$ & $77.3(\pm 3.8)$ & $77.8(\pm 3.2)$ & $78.3(\pm 2.8)$ & $76.9(\pm 3.7)$ & $78.2(\pm 3.3)$ & $\bf 78.6(\pm 3.1)$ \\ \cline{3-10} 
                  &                   &  $\forget$& $\bf 0.0 (\pm 0.0)$ & $3.2 (\pm 0.6)$ & $3.7 (\pm 0.5)$ & $\bf 0.0 (\pm 0.0)$ & $5.8 (\pm 0.6)$ & $0.0 (\pm 0.0)$ & $\bf 0.0 (\pm 0.0)$ \\ \cline{2-10} 
                  & \multirow{2}{*}{20} & $\remain$ & $\bf 78.6 (\pm 3.2)$ & $76.3 (\pm 2.8)$ & $76.9 (\pm 2.8)$ & $76.6 (\pm 2.6)$ & $75.1 (\pm 2.6)$ & $76.9 (\pm 2.7)$ & $\bf 77.3 (\pm 2.8)$ \\ \cline{3-10} 
                  &                   &$\forget$  & $\bf 0.0 (\pm 0.0)$ & $8.7 (\pm 0.9)$ & $8.2 (\pm 0.8)$ & $7.6 (\pm 0.6)$ & $9.2 (\pm 0.9)$ & $8.3 (\pm 0.6)$ & $\bf 7.5 (\pm 0.7)$ \\ \hline
\multirow{4}{*}{\thead{RES34+ \\CUB200}} & \multirow{2}{*}{1} & $\remain$ & $\bf 85.1 (\pm 4.1)$ & $83.5 (\pm 3.8)$ & $83.7 (\pm 3.8)$ & $83.9 (\pm 3.6)$ & $83.4 (\pm 4.1)$ & $83.7 (\pm 3.9)$ & $\bf 84.2 (\pm 3.8)$ \\ \cline{3-10} 
                  &                   & $\forget$ & $\bf 0.0 (\pm 0.0)$ & $3.5 (\pm 0.8)$ & $4.1 (\pm 0.6)$ & $\bf 0.0 (\pm 0.0)$ & $4.3 (\pm 0.9)$ & $\bf 0.0 (\pm 0.0)$ & $\bf 0.0 (\pm 0.0)$ \\ \cline{2-10} 
                  & \multirow{2}{*}{20} &  $\remain$& $\bf 84.9 (\pm 3.7)$ & $82.6 (\pm 3.5)$ & $82.5 (\pm 3.6)$ & $\bf 83.5 (\pm 3.3)$ & $82.7 (\pm 3.9)$ & $82.9 (\pm 3.6)$ & $\bf 83.5 (\pm 3.6)$ \\ \cline{3-10} 
                  &                   &  $\forget$& $\bf 0.0 (\pm 0.0)$ & $9.7 (\pm 0.7)$ & $8.9 (\pm 0.6)$ & $8.3 (\pm 0.5)$ & $10.2 (\pm 0.7)$ & $8.5 (\pm 0.6)$ & $\bf 7.9 (\pm 0.7)$ \\
\bottomrule
\end{tabular}
}
\vspace{-0.5cm}
\end{table}
we evaluate the unlearning model's effectiveness under full-class (single and multiple classes) deletion. Specifically, we randomly selected either 1 or 20 classes and removed all associated samples, repeating this process five times for reliable results. The experiments were conducted using CIFAR100 and CUB200 datasets with ResNet18 and ResNet34 architectures, updating the model for one epoch at a learning rate of 0.001. Our objective was to diminish the model's accuracy for the removed ("forget") classes while maintaining accuracy for the remaining ("remain") classes. One can see from Tab.~\ref{tab:class-unlearning} that the retrained model shows the ground-truth accuracy on $\remain$ and zero for $\forget$ due to the classes' complete removal.  Notably, while the SCRUB method showed decent performance in single class deletions, it struggled in scenarios involving multiple class deletions, securing only 74.9\% accuracy for "remain" and 6.3\% for "forget", attributed to its random class assignment for forgotten samples. Our methodology surpasses all compared baselines in accuracy metrics, benefiting from a debiased unlearning approach that ensures robust performance across both single and multiple-class deletions. Specifically, our technique exceeds SCRUB by 3\% in RA and 13\% in FA in the CIFAR-100 scenario with 20 classes removed.

\subsection{Evaluate Model Resilience to Unlearning Bias under a Non-uniform Deletion Strategy(Q2)}
\subsubsection{Selective Class Deletion}
\begin{figure}[t]
\centering
\includegraphics[width=1\columnwidth]{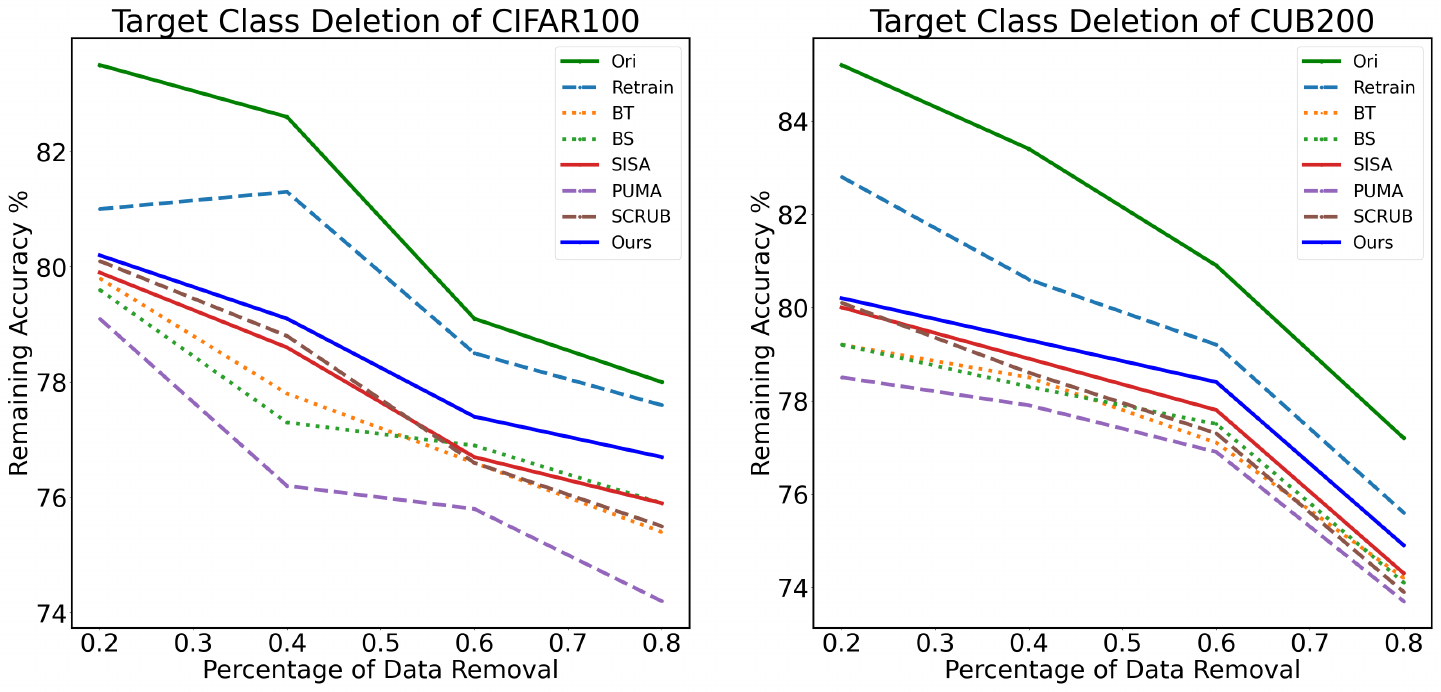}
\vspace{-0.2cm}
\caption{The RA of Targeted Class Deletion with Different Percentage of Data Removal.}
\label{fig:RA}
\vspace{-0.6cm}

\end{figure}

We evaluate the performance of unlearning models in scenarios where data is selectively forgotten, leading to label bias. We simulate imbalanced datasets by removing 20\%, 40\%, 60\%, and 80\% of samples from specific classes—namely, "Rabbit" in CIFAR100 and "Cardinal" in CUB200—and analyze the effectiveness of our method, compared to others, in mitigating this bias by \textit{RA}.
The results (see Fig~\ref{fig:RA}) show that our approach consistently outperforms baseline methods (SCRUB and SISA) across all levels of data deletion. While SCRUB and SISA maintain satisfactory performance at 20\% data removal, their effectiveness drops markedly at higher removal percentages. Specifically, when removing 60\% data, SCRUB misclassifies 72\% of the remaining "Rabbit" instances, compared to our method's lower misclassification rate of 33\%. This highlights the superior robustness of our method against increasing levels of label bias due to biased data deletion.

\subsubsection{Selective Feature Deletion}
We evaluate our approach in scenarios where data deletion is biased towards specific attributes, leading to a shortcut bias. Here, we adopt two datasets, \textit{Adult} and \textit{German} with favourable labels as \textit{high salary} ($\hat{y}=1$) and \textit{good credit risks} ($\hat{y}=1$) respectively. Following~\cite{zhang2024forgotten}, in both datasets, we define \textit{sex} as a sensitive feature, categorizing females as the unprivileged group. We assume that $30\%$ people from unprivileged groups request data removal. To compare the bias mitigation, we conduct the non-uniform deletion and measure the \textbf{\textit{DI}}, \textbf{\textit{EOD}} as well as the \textbf{\textit{RA}} and \textbf{\textit{MIA}}.  
Initially, we develop an \textit{unbiased} baseline model (Ori), following \cite{oesterling2023fair}. Then, we selectively remove instances with specific features from the model. Our results indicate that our method closely matches the performance of an unbiased retraining approach, achieving a \textit{DI} of 0.85 for \textit{Adult} and 0.89 for \textit{German}. Although \textit{F-unlearning} reduces bias by adjusting its fairness regularization parameter, it significantly decreases utility, resulting in a low accuracy on $\remain$. This is because directly minimizing the regularizer adversely affects the cross-entropy loss. Our model showcases its superior ability to address shortcut bias inherent in the unlearning process.

\begin{table}[h!]
\centering
\caption{Result under Non-uniform Deletion Strategy.}
\label{tab:Nonuniform}
\scalebox{0.9}{
\begin{tabular}{c|cccc|cccc}
\toprule
\textbf{Dataset ($\rightarrow$)} & \multicolumn{4}{c|}{\textbf{Adult}}                                   & \multicolumn{4}{c}{\textbf{German}}                                \\
\hline
{\textbf{Algorithm} ($\downarrow$)}{\textbf{Metrics} ($\rightarrow$)}
        & \multicolumn{1}{c|}{RA}   & \multicolumn{1}{c|}{MIA}  & \multicolumn{1}{c|}{DI}  & EOD & \multicolumn{1}{c|}{RA}   & \multicolumn{1}{c|}{MIA}  & \multicolumn{1}{c|}{DI} & EOD \\
\midrule
Ori     & \multicolumn{1}{l|}{$\bf 0.82$} & \multicolumn{1}{l|}{$NA$}  &\multicolumn{1}{l|}{$\bf 0.93$} & \multicolumn{1}{l|}{$\bf -0.02$} & \multicolumn{1}{l|}{$\bf 0.73$} & \multicolumn{1}{l|}{$NA$} & \multicolumn{1}{l|}{$\bf 0.95$} & $\bf -0.01$  \\ \hline
Retrain    & \multicolumn{1}{l|}{$\bf 0.76$} & \multicolumn{1}{l|}{$\bf 0.02$} &\multicolumn{1}{l|}{$\bf 0.89$} & \multicolumn{1}{l|}{$\bf -0.03$} & \multicolumn{1}{l|}{$\bf 0.68$} & \multicolumn{1}{l|}{$\bf 0.04$} & \multicolumn{1}{l|}{$\bf 0.91$} &$\bf -0.01$  \\ \hline
BT      & \multicolumn{1}{l|}{$0.70$} & \multicolumn{1}{l|}{$0.06$} &\multicolumn{1}{l|}{$0.76$} & \multicolumn{1}{l|}{$-0.09$} & \multicolumn{1}{l|}{$0.62$} &  \multicolumn{1}{l|}{$0.08$} & \multicolumn{1}{l|}{$0.84$} & $-0.06$  \\ \hline
BS      & \multicolumn{1}{l|}{$\bf 0.74$} & \multicolumn{1}{l|}{0.03} &\multicolumn{1}{l|}{$0.77$} & \multicolumn{1}{l|}{$-0.07$} & \multicolumn{1}{l|}{$0.63$} &  \multicolumn{1}{l|}{$0.06$} & \multicolumn{1}{l|}{$0.84$} &$-0.06$  \\ \hline
PUMA   & \multicolumn{1}{l|}{0.69} & \multicolumn{1}{l|}{$0.06$} &\multicolumn{1}{l|}{0.73} & \multicolumn{1}{l|}{$-0.07$} & \multicolumn{1}{l|}{$0.62$} & \multicolumn{1}{l|}{$0.09$} & \multicolumn{1}{l|}{$0.81$} &$-0.06$  \\ \hline
SCRUB    & \multicolumn{1}{l|}{$0.72$} & \multicolumn{1}{l|}{$\bf 0.03$} &\multicolumn{1}{l|}{$0.76$} & \multicolumn{1}{l|}{$-0.06$} & \multicolumn{1}{l|}{$\bf 0.65$} & \multicolumn{1}{l|}{$0.06$} & \multicolumn{1}{l|}{$0.82$} &$-0.07$  \\ \hline
F-Unlearning    & \multicolumn{1}{l|}{$0.68$} & \multicolumn{1}{l|}{$0.04$} &\multicolumn{1}{l|}{$0.79$} & \multicolumn{1}{l|}{$-0.05$} & \multicolumn{1}{l|}{$0.60$} & \multicolumn{1}{l|}{$\bf 0.05$} & \multicolumn{1}{l|}{$0.84$} &$-0.07$  \\ \hline
Ours    & \multicolumn{1}{l|}{$\bf 0.74$} & \multicolumn{1}{l|}{$\bf 0.03$} &\multicolumn{1}{l|}{$\bf 0.85$} & \multicolumn{1}{l|}{$\bf -0.03$} & \multicolumn{1}{l|}{$\bf 0.65$} & \multicolumn{1}{l|}{$\bf 0.05$} & \multicolumn{1}{l|}{$\bf 0.89$} & $\bf -0.03$   \\
\bottomrule
\end{tabular}
}

\end{table}
\vspace{-0.5cm}

\subsection{Ablation Studies on Loss Function (Q3)}
We examine the effects of each component of the loss function, by ablating each part from $\mathcal{L}$. Notice that $\Loss_r$ is consistently retained within $\mathcal{L}$ to guarantee utility on $\remain$. The ablation study is conducted on CIFAR100 with ResNet18 by uniformly removing 10\% and 40\% samples. In the full model, we set $\alpha=0.6,\beta=0.2,\gamma=0.2$. 

\begin{table}[]
\centering
\setlength{\tabcolsep}{2pt}
\caption{Performance of removing one component of the loss function from $\mathcal{L}$.}
\label{tab:Ablation}
\begin{tabular}{c|cc|cc}
\toprule
 & \multicolumn{2}{c|}{{\bf 10\% Data Removal}}  & \multicolumn{2}{c}{{\bf 40\% Data Removal}} \\
 \cmidrule(lr){2-5}
 & RA & MIA & RA & MIA \\
 \midrule
 \multicolumn{1}{c|}{$\alpha=0$} & $0.79 (\pm 0.03)$ & $0.04 (\pm 0.003)$ & $0.67 (\pm 0.02)$ & $0.15 (\pm 0.003)$ \\
 \multicolumn{1}{c|}{$\beta=0$} & $0.83 (\pm 0.03)$ & $0.07 (\pm 0.005)$ & $0.69(\pm 0.04)$ & $0.17 (\pm 0.004)$ \\
 \multicolumn{1}{c|}{$\gamma=0$} & $\bf 0.89(\pm 0.02)$ & $\bf 0.03 (\pm 0.007)$ & $\bf 0.77 (\pm 0.03)$ & $\bf 0.09 (\pm 0.003)$ \\
  \multicolumn{1}{c|}{$\alpha=0, \beta=0$} & $0.73 (\pm 0.04)$ & $0.15 (\pm 0.003)$ & $0.61 (\pm 0.03)$ & $0.24 (\pm 0.003)$ \\
   \multicolumn{1}{c|}{$\alpha=0, \gamma=0$} & $0.76 (\pm 0.05)$ & $\bf 0.03 (\pm 0.004)$ & $0.63 (\pm 0.04)$ & $0.12(\pm 0.006)$ \\
   \multicolumn{1}{c|}{$\beta=0, \gamma=0$} & $0.80 (\pm 0.03)$ & $0.09 (\pm 0.004)$ & $0.70 (\pm 0.02)$ & $0.17 (\pm 0.004)$ \\
 \multicolumn{1}{c|}{Full} & $\bf 0.90 (\pm 0.02)$ & $\bf 0.02 (\pm 0.005)$ & $\bf 0.79 (\pm 0.02)$ & $\bf 0.09 (\pm 0.002)$ \\
 
\bottomrule
\end{tabular}
\end{table}

Comparing the performance drop of each component, we observe that $\Loss_{f}$ crucially impacts the balance between remembering and forgetting. Table~\ref{tab:Ablation} shows that keeping only $\Loss_{f}$ within $\mathcal{L}$ leads to good performance with an accuracy of $0.8$, compared to the full model's $0.9$. However, it is not as effective in forgetting information as $\Loss_{f}^{CB}$, which achieves a MIA rate slightly worse than the full model. This is because $\Loss_{f}^{CB}$ directly reduces the geometric distance between forgotten samples and their corresponding CF in representation space, intentionally guiding the model to overlook certain local features. Hence, combining $\Loss_{f}$ and $\Loss_{f}^{CB}$ outperforms other settings. Finally, $\ell(X_i, Y_i)$ contributes to maintaining the accuracy of the remaining samples.

\section{Conclusion and Future Work}
\label{sec:conclusion}

In this work, we tackled the problem of mitigating the bias introduced by machine unlearning techniques~\cite{xie2022measurement, chen2022multi}. 
To achieve this goal, we proposed a causal framework that addresses data-level bias through causal intervention~\cite{lyu2023attention, mao2022trace}.
Furthermore, we guided the forgetting procedure using counterfactual examples that mitigate the algorithmic bias maintaining semantic data consistency without impacting the remaining samples~\cite{zhang2023unleashing, xie2021vitalhub,lahn2024combinatorial}.
Experimental results demonstrated that our approach outperforms existing machine unlearning baseline on standard quality metrics. In the future, we plan to extend our work on NLP tasks~\cite{lyu2022study}.

\small
\bibliographystyle{plain}

 \end{document}